\documentclass{llncs}
\usepackage[utf8]{inputenc}
\usepackage[scaled=0.85]{beramono}
\usepackage{listings}
\usepackage{paralist}
\usepackage{parcolumns}
\usepackage{multicol}
\usepackage{graphicx}
\usepackage{mathtools}
\usepackage[linesnumbered,ruled]{algorithm2e}
\usepackage{algpseudocode}
\usepackage{amsfonts}
\usepackage{amssymb}
\usepackage{verbatim}
\usepackage{standalone}
\usepackage{url}
\usepackage{subfig}
\usepackage{float}
\usepackage[table,xcdraw]{xcolor}
\usepackage{hyperref}
\usepackage{cleveref}
\usepackage{caption} 
\usepackage{amssymb}
\usepackage{todonotes}

\usepackage{soul}
\usepackage{paralist}

\begin{document}

\title{Feature-based Reformulation of Entities in Triple Pattern Queries}
\titlerunning{Reformulating Entity Triple Patterns}  
%
\author{Amar Viswanathan\inst{1} \and Geeth de Mel\inst{2} \and 
James A.Hendler\inst{1}}
\authorrunning{Viswanathan et al.} 
%
\tocauthor{Amar Viswanathan, Geeth de Mel, James A.Hendler}
\institute{Rensselaer Polytechnic Institute, Troy NY 12180, USA,\\
\email{kannaa@rpi.edu,hendler@cs.rpi.edu}\\
\and
IBM Research, Warrington, UK\\
\email{geeth.demel@uk.ibm.com}}

\maketitle              

\begin{abstract}

Knowledge graphs encode uniquely identifiable entities to other entities or literal values by means of relationships, thus enabling semantically rich querying over the stored data. Typically, the semantics of such queries are often crisp thereby resulting in crisp answers. Query log statistics show that a majority of the queries issued to knowledge graphs are often entity centric queries. When a user needs additional answers the state-of-the-art in assisting users is to rewrite the original query resulting in a set of approximations. Several strategies have been proposed in past to address this. They typically move up the taxonomy to relax a specific element to a more generic element. Entities don't have a taxonomy and they end up being generalized. To address this issue, in this paper, we propose an entity centric reformulation strategy that utilizes schema information and entity features present in the graph to suggest rewrites. Once the features are identified, the entity in concern is reformulated as a set of features. Since entities can have a large number of features, we introduce strategies that select the top-k most \emph{relevant} and \emph{informative} ranked features and augment them to the original query to create a valid reformulation.  We then evaluate our approach by showing that our reformulation strategy produces results that are more informative when compared with state-of-the-art.

\end{abstract}
\section{Introduction} 
Knowledge graphs link uniquely identifiable entity nodes  to other such entity nodes or literal values by means of relationship edges; this is achieved by means of \emph{class-} and \emph{predicate- hierarchies} encoded as schemata.  Therefore, every unique node-edge-node in the graph can be represented as a \emph{Subject-Predicate-Object} ($\mathbb{SPO}$) triple where object refers to an entity or a literal. This is the foundation for the Resource Description Framework (RDF)~\cite{schreiber2014rdf}, which is the keystone for the Semantic Web; popular knowledge graphs like YAGO, DBpedia and Freebase~\cite{yago,dbpedia,freebase} use variations of this representation to capture the semantic meaning of the content published by user communities. RDF based knowledge graphs are stored in \emph{triple stores} and the information stored is queried using \emph{triple pattern queries}.


The semantics behind knowledge graphs enable the use of complex conditions in an intuitive manner while querying for information which adds to the allure of knowledge graphs.  Thus, knowledge graphs are used quite often in data intensive artificial intelligence (AI) applications where data storage efficiencies and context of operation are of paramount importance. However, formulating queries that translate the user intent to the right answer is not trivial.  This is due to the fact that large knowledge graphs are created automatically and change over time as the conceptual understanding of the domain increases. Thus, the onus is on the user to understand the underlying schema and instance data distributions to ask the right questions; this leads to manual trial and error processes in which users reformulates and rewrites their original query until the information needs are met which frustrates the user. 

\begin{figure}[htp]
\centering
\includegraphics[scale=0.5]{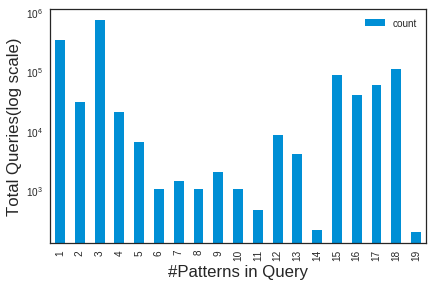}
\caption{Distribution of Entity Queries in the Linked SPARQL Dataset}
\label{fig:queryDistribution}
\end{figure}

Flexible querying systems have addressed this issue by suggesting \emph{automatic query reformulations}, where the reformulations are approximately similar to the original query. These reformulations are typically generated by utilizing the underlying ontology hierarchies---i.e., class- and property- hierarchies of the knowledge graph, especially in situations where additional answers are required~\cite{hurtado2006,hurtado2008}. By traversing up the hierarchy, the restrictive input patterns in the original query are loosened or relaxed to yield more answers. This works well for queries that contain schema information such as \emph{classes} and \emph{properties}. However, when the query patterns contain \emph{entities}---i.e., instantiations of classes---
they are relaxed to a variable. This is known as \textit{simple relaxation}. \cite{hurtado2006,hurtado2008,fokou2015,fokou2016,huang2012}. However, once a simple relaxation is applied to an entity, all the semantics associated with the entity is also disregarded which is a huge drawback.  The problem is exacerbated even more by the observation that according to the Linked SPARQL Dataset(LSQ)~\cite{saleem2015lsq}, majority of graph queries to public endpoints are of the form entity queries as shown by Figure~\ref{fig:queryDistribution}. 

\vspace{-0.5cm}
\vspace{-0.5cm}

\begin{center}
\begin{figure}[htp]
\fbox{\includegraphics[width=\linewidth]{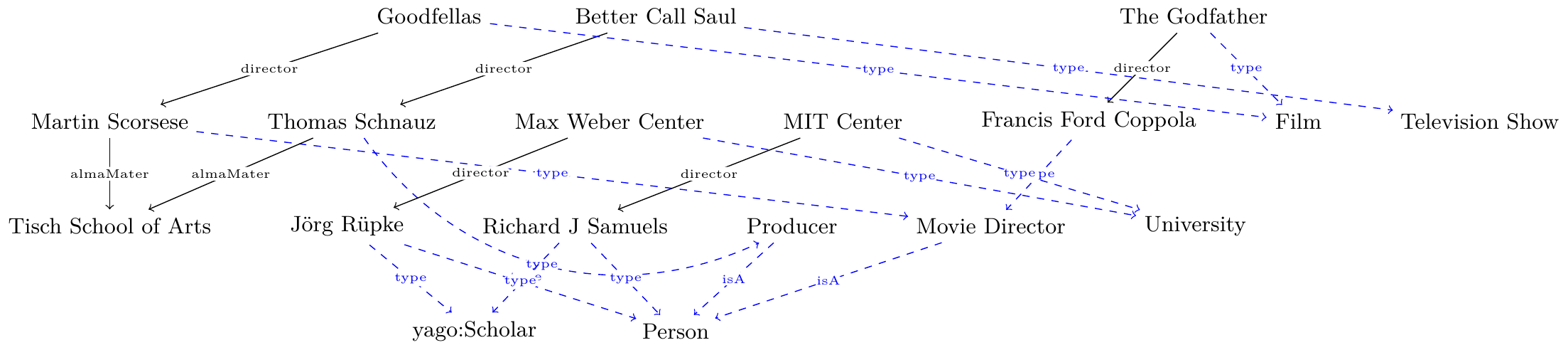}}
\caption{Knowledge Graph Snippet}
\label{fig:KnowledgeGraph}
\end{figure}
\end{center}

Motivated by this observation, we argue that it would be beneficial to look at approaches which preserve the semantics of the entities while reformulating queries. To illustrate the need further, let us look at~\autoref{fig:KnowledgeGraph}, which shows a snippet of a knowledge graph adapted from \href{http://downloads.dbpedia.org/current/}{DBpedia 2015-10}. We assume the initial query to be \emph{Anything directed by Martin\_Scorsese}. Now \autoref{tab:KG-rel-table} shows the effect of applying simple relaxation on this query---i.e., simple relaxation is applied to the instance \emph{Martin\_Scorsese} to change it to a variable \textbf{?o}. Now this translates as \emph{Anything directed by}~\textbf{?o}.  This results in answers that include organizations such as \emph{Max\_Weber\_Center}, \emph{MIT\_Center}, and so forth. A glance at the graph in Figure~\ref{fig:KnowledgeGraph} shows this is \emph{logically correct} but conceptually not what the user intended. If relaxed further this would lead to selecting the entire graph. Some of the potential drawbacks of this are as follows:
\begin{itemize}
\item the semantics of instance data is completely ignored---i.e., $\mathbb{SPO}$ statements of the entity are not used for relaxation;
\item the relaxed version is a generalization, thus leads to high recall in large graphs;
\item a single pattern in the input query can end up selecting the entire graph, which results in higher query costs; and 
\item more than one query or query pattern can lead to the same generalization, thus resulting in the same insight for vastly different queries. 
\end{itemize}

\vspace{-0.5cm}

\vspace{-1.5cm}

\begin{table}[htp]
\centering
\scalebox{0.85}{
\setlength{\tabcolsep}{10pt}
\begin{tabular}{llc}
\multicolumn{1}{c}{\textbf{Initial Query}}               & \multicolumn{1}{c}{\textbf{Relaxation \#1}}                                                                                                   & \textbf{Relaxation \#2}                        \\ \hline
\multicolumn{1}{|l|}{?s :director :Martin\_Scorsese} & \multicolumn{1}{l|}{?s :director \textbf{?o}.}                                                                                          & \multicolumn{1}{l|}{?s \textbf{?p ?o} .} \\ \hline
\multicolumn{1}{|c|}{\textbf{Results}}                   & \multicolumn{1}{c|}{\textbf{Results}}                                                                                                         & \multicolumn{1}{c|}{\textbf{Results}}          \\ \hline
\multicolumn{1}{|l|}{The\_Godfather}                      & \multicolumn{1}{l|}{\begin{tabular}[c]{@{}l@{}}The\_Godfather\\ Goodfellas\\ Better\_Call \_Saul\\ Max\_Weber\_Center\\ MIT\_Center\end{tabular}} & \multicolumn{1}{c|}{ $\infty$}                       \\ \hline
\end{tabular}
}
\caption{Relaxation and Results}
\label{tab:KG-rel-table}
\end{table}
\vspace{-0.5cm}

Given the query \emph{Anything directed by Martin\_Scorsese}, it would be beneficial to suggest a reformulation that yields entities related to \textit{Martin\_Scorsese} and then relax gradually. For example, the query \emph{Anything directed by people who attended Tisch\_School\_of\_the\_Arts} is a more contextual reformulation than relaxing to a variable. 
Therefore, we propose two modifications to existing techniques that add more context to the query. First, we contextualize the domain of values that the variables in the entity triple pattern can assume. Next, we unfold the entity as an entity pattern utilizing the instance data statements. To the best of our knowledge, no work exists on entity reformulation by utilizing its \textit{instance data statements} for approximate answers. We summarize our contributions as follows:
\begin{itemize}
\item propose an approach for query reformulation as an entity reformulation problem;  
\item a set of rules to reformulate the entities using the $\mathbb{SPO}$ statements in the graph; and 
\item means to rank the top-k most \emph{relevant} and \emph{informative} features of an entity by defining an information retrieval (IR) inspired feature ranking mechanism. In large graphs, we use this to pick the best possible reformulation.
\end{itemize}

The rest of the paper is organized as follows: Section~\ref{sec:preliminaries} details the syntax and the semantics of the vocabulary used for the paper. Section~\ref{sec:reformulation} presents our approach to reformulate queries w.r.t. triple patterns; since large knowledge graphs can have a high number of entity statements, Section~\ref{sec:reformulation} also deals with summarizing entity statements by utilizing a ranking mechanism. We then evaluate the model in Section~\ref{sec:evaluation} by comparing it against some well known simple relaxation mechanisms and discuss the applicability of our work in real-world situations. Finally, we discuss the related work in Section~\ref{sec:discussion} and conclude the paper in Section~\ref{sec:conclusion} by sketching directions for future research.

\section{Preliminaries}~\label{sec:preliminaries}
In this section we will introduce the syntax and the semantics of the knowledge graph, triple patterns, and triple pattern queries we will use throughout the paper.

\subsection{Data Model}
 Assuming that there is an infinite set of IRIs $\mathbb{I}$, 
Literals $\mathbb{L}$ and blank nodes $\mathbb{B}$, then the elements in $\mathbb{I} \cup \mathbb{B} \cup \mathbb{L}$ are RDF terms. They are represented as a set of $\langle v_1,v_2,v_3 \rangle \in \left( \mathbb{I} \cup \mathbb{B}\right) \times  \mathbb{I} \times \left( \mathbb{I} \cup \mathbb{B} \cup \mathbb{L} \right)$ \textit{triple} statements where $v_1,v_2,v_3$ are called as $\langle subject, predicate, object \rangle$ \cite{hurtado2006}. The \textit{subject} values can also describe entities $\mathbb{E}$, which in addition have \textit{type statements} that group them to classes $\mathbb{C}$ and \textit{schema statements} that group classes to their super classes. These statements are generally part of what is known as an \textit{Ontology} or schema. 

\subsection{Triple Pattern} 
The building block of all knowledge graph query languages is a triple pattern. Given a set of Variables $\mathbb{V}$, which are distinct from $\mathbb{I}, \mathbb{L}, \mathbb{E}, \mathbb{C}$, a triple pattern \textit{t} $ \in \left( \mathbb{V} \cup \mathbb{I} \right) \times  \mathbb{I} \cup \mathbb{V} \times \left( \mathbb{V} \cup \mathbb{I}  \cup  \mathbb{L} \right)$ has a variable prefixed with a question mark and can be placed either at the subject, predicate, object or any two or all three in a triple statement; example formation of triple patterns are  $\left\lbrace \left\langle ?s, p,o \right\rangle, \left\langle s, ?p,o \right\rangle, \ldots
\left\langle ?s, ?p,?o \right\rangle \right\rbrace$. 


\subsection{Triple Pattern Query}   
A triple pattern query or a basic graph pattern (BGP) query Q is an expression of the form H $\leftarrow \mathbb{B}$, where the body $\mathbb{B}$ is a set of triple patterns $\left\lbrace t_1, t_2, \ldots t_n \right\rbrace$ and the head H contains a set of variables belonging to $\mathbb{B}$. This definition can easily be extended to any type of queries as stated by Hurtado et al.\cite{hurtado2006}. A matching function $\Theta$ is defined from the variables in \textbf{B} to the Blanks, IRIs and literals in the graph G. Thus, $\Theta(\mathbb{B}(Q))$ is a graph obtained by replacing each variable X in the body by $\Theta(X)$. Given an RDF graph G, the \textit{solution} of the BGP query Q is defined as: for each matching $\Theta$ such that $\Theta(B(Q)) \subseteq cl(G)$, return $\Theta(H(Q))$, where cl(G) is the \textit{closure} of the graph G\cite{hayes}. \autoref{tab:KG-rel-table} shows examples of triple pattern queries and the answers when the query is executed on the graph \autoref{fig:KnowledgeGraph}


In the next section, we shall introduce our query reformulation strategy.

\section{Triple Pattern Reformulation Model}\label{sec:reformulation}~\label{sec:triple-ref}
Given a triple pattern, our reformulation model comprises of a set of relaxation rules for the class and property elements. We then propose rules for the \textit{constants} in the triple pattern. We extend the definition of constants described in Hurtado \textit{et al.}~\cite{hurtado2006} and Fokou~\textit{et al.}~\cite{fokou2016} to literal values and entities or instances~\footnote{datatype properties link to literals and object type properties link to instances}. In addition we also define rules to assign \emph{type} information to a \textit{elements} present in the entity triple patterns. While Hurtado \textit{et al.} replaced this triple, we suggest appending a triple pattern to the existing triple pattern, thus preserving semantics related to the original entity. 
\begin{itemize}
\item \textit{Superclass Relaxation} : $\langle s, type, c_1 \rangle  \Rightarrow$ $\langle s, type, c_2 \rangle$, if $c_1$ subClassOf $c_2$. 
\item \textit{Superproperty Relaxation} :  $\langle s, p_1, o \rangle  \Rightarrow$ $\langle s, p_2, o \rangle$, if $p_1$ subPropertyOf $p_2$.
\item \textit{Literal Relaxation}: If \textit{l} is a literal value present in a triple \textit{t}, then \textit{t} $\Rightarrow$ \textit{t'} by replacing \textit{l} by a variable. Here \textit{t'} is the relaxed triple pattern.
\item \textit{Variable Typing}:$\langle ?v,p,e \rangle \Rightarrow$ $( \langle ?v, type, c_1 \rangle, \langle ?v,type, c_2 \rangle, \ldots, \langle ?v, type, c_k \rangle,$ \\$\langle ?v, p, e \rangle )$, where $c_1, c_2, \ldots, c_k$ are the types for variables i.e. $\mid type(?v) \mid ?v,p \in \langle ?v,p,o \rangle \mid$.
\item \textit{Entity Reformulation}: For entity \textit{e} present in a triple pattern and there exists triples $\langle e_1,p_1,o_1 \rangle, \ldots, \langle e,p_k,o_k \rangle$ for \textit{e} in the \textit{cl}(G), the entity reformulation rules are shown in Table.2:
\end{itemize}

\begin{table}[htp]
\centering
\scalebox{0.85}{
\begin{tabular}{l|l|l|}
\cline{2-3}
                         & \multicolumn{1}{c|}{\textbf{Triple Pattern}} & \multicolumn{1}{c|}{\textbf{Entity Reformulation}}                                                                                                                                                                                                                                                                                                                                                                        \\ \hline
\multicolumn{1}{|l|}{1.} & $\left\langle e,?p,?o \right\rangle$         & $ \langle ?e,?p,?o \rangle$                                                                                                                                                                                                                                                                       \\ \hline
\multicolumn{1}{|l|}{2.} & $\left\langle e,P,?o \right\rangle$          & \begin{tabular}[c]{@{}l@{}}$\left\lbrace \left\langle ?e,P,?o \right\rangle,\left\langle ?e,p_1,o_1 \right\rangle,\left\langle ?e,p_2,o_2 \right\rangle, \ldots,\left\langle ?e,p_k,o_k \right\rangle\right\rbrace$\\ \\ $P \notin \left\lbrace p_1,p_2, \ldots p_k \right\rbrace$\end{tabular}                                                                                                                              \\ \hline
\multicolumn{1}{|l|}{3.} & $\left\langle e,P,o \right\rangle$          & \begin{tabular}[c]{@{}l@{}}$\left\lbrace \left\langle ?e,P,o \right\rangle,\left\langle ?e,p_1,o_1 \right\rangle,\left\langle ?e,p_2,o_2 \right\rangle, \ldots,\left\langle ?e,p_k,o_k \right\rangle\right\rbrace$\\ \\ $\left\langle P,o \right\rangle \notin \left\lbrace \left\langle p_1, o_1 \right\rangle, \left\langle p_2, o_2 \right\rangle, \ldots, \left\langle p_k, o_k\right\rangle \right\rbrace$\end{tabular} \\ \hline
\multicolumn{1}{|l|}{4.} & $\left\langle s,p,e\right\rangle$            & \begin{tabular}[c]{@{}l@{}}$\left\lbrace \left\langle s,p,?e \right\rangle,\left\langle ?e,p_1,o_1 \right\rangle,\left\langle ?e,p_2,o_2 \right\rangle, \ldots,\left\langle ?e,p_k,o_k \right\rangle\right\rbrace$\\ \\ $p \notin \left\lbrace p_1,p_2, \ldots p_k \right\rbrace$\end{tabular}                                                                                                                           \\ \hline
\multicolumn{1}{|l|}{5.} & $\left\langle ?   s,p,e\right\rangle$        & \begin{tabular}[c]{@{}l@{}}$\left\lbrace \left\langle ?s,p,?e \right\rangle\left\langle ?e,p_1,o_1 \right\rangle,\left\langle ?e,p_2,o_2 \right\rangle, \ldots,\left\langle ?e,p_k,o_k \right\rangle\right\rbrace$\\ \\ $p \notin \left\lbrace p_1,p_2, \ldots p_k \right\rbrace$\end{tabular}                                                                                                                          \\ \hline
\multicolumn{1}{|l|}{6.} & $\left\langle ?s,?p,e\right\rangle$          & $\left\lbrace \left\langle ?s,?p,?e \right\rangle,\left\langle ?e,p_1,o_1 \right\rangle\left\langle ?e,p_2,o_2 \right\rangle, \ldots,\left\langle ?e,p_k,o_k \right\rangle\right\rbrace$                                                                                                                                                                                                                                \\ \hline
\end{tabular}
}
\caption{Entity Reformulation Rules}
\label{tab:rules}
\end{table}
\vspace{-0.2cm}
Given a triple $\langle s, p, e \rangle$ where \textit{e} is an entity resource with statements $\langle e,p_1,o_1 \rangle,$ \\$\ldots, \langle e,p_k,o_k \rangle$, executing the  pattern $\langle s,p,e \rangle$ is equal to executing the pattern
$\langle s,p,?e \rangle,\langle ?e,p_1,o_1 \rangle, \ldots, \langle ?e,p_k,o_k \rangle$. This is because the variable \textit{?e} will be bounded to the entity \textit{e}. Utilizing this notion \autoref{tab:rules} shows all the rules for rewriting an entity in the triple pattern. Except for the first rule, where the triple pattern selects all the statements in the graph, all the other rules utilize the entity statements in the graph and append them back to the query. For smaller graphs like LUBM\cite{guo2005lubm}, the number of entity properties are less and so appending the top-\textit{k} does not make the reformulated query pattern large. However, knowledge graphs such as DBpedia \cite{dbpedia} have a lot of facts per entity and appending them back to the original query would increase the execution time of a query. In addition to this, it would not be meaningful to the user when a reformulation with a lot of conditions is presented. Thus, we present a technique that ranks the facts according to their fact importance and then selects the top-\textit{k} facts and converts them to \textit{entity summary patterns}. These top-\textit{k} patterns are then appended to the query. Thus we define our problem statement as :
\subsubsection{Problem Statement} Given an entity in the triple pattern and the entity reformulation rules, we are concerned with picking the top-\textit{k} most relevant patterns so that the reformulations that are more \textit{contextual} and \textit{specific} than a relaxation.
\subsection{Entity Fact Summarization}
\subsubsection{Definition 3} \textbf{Entity Fact} A fact \textit{f$_i$} of an entity $e \in \mathbb{E}$ is the $\langle p, o \rangle$ pair $\in \left( \mathbb{U} \right) \times  \left( \mathbb{U} \cup  \mathbb{L} \right)$ in the triple statement $\langle e,p,o \rangle$ for the input graph G. In this paper we focus only on those facts whose object values are resources or entities. In future work, we will be taking upon literal values as well.
\subsubsection{Definition 4} \textbf{Entity Fact Set} Entities have many facts associated with them in the knowledge graph and this is denoted by the fact set \textit{FS(e)}, which is the total number of facts for the entity e in G.

\subsubsection{Entity Summary}
Since the number of facts for an entity can be quite high in large knowledge graphs, we define a ranked summary of such facts. Given an entity \textit{e} and a positive integer  \textit{k} $\leq \mid\textit{FS(e)}\mid$, an entity summary of \textit{e}, $\textit{Summ(e,k)} \subseteq \textit{FS(e)}$ such that $\mid\textit{Summ(e,k)}\mid = k$. The facts $\left\lbrace f_1,f_2,\ldots f_k \right\rbrace $ in the fact set \textit{FS(e)} are ranked with ranking function such that $Rank(f_1) > Rank(f_2) > \ldots Rank(f_k)$ 

\vspace{-0.5cm}
\begin{table}[H]
\centering
\scalebox{0.75}{
\setlength{\tabcolsep}{5pt}
\begin{tabular}{|c|c|c|}
\hline
\textbf{Entity Fact Set}                                                                                                                                                                                                                              & \textbf{Entity Summary}                                                                                                                                & \textbf{Entity Summary Pattern}                                                                                                                                 \\ \hline
\textbf{:Martin\_Scorsese}                                                                                                                                                                                                                            & \textbf{:Martin\_Scorsese}                                                                                                                             & \textbf{?s}                                                                                                                                                     \\ \hline
\textbf{\textless p,o\textgreater}                                                                                                                                                                                                                     & \textbf{\textless p,o\textgreater, k = 3}                                                                                                               & \textbf{\textless?s,p,o\textgreater}                                                                                                                            \\ \hline
\multicolumn{1}{|l|}{\begin{tabular}[c]{@{}l@{}}:almaMater :New\_York\_University\\ :birthPlace :Queens\\ :spouse :Isabella\_Rossellini\\ :parents :Catherine\_Scorsese\\ :parents :Charles\_Scorsese\\ :subject :Tisch\_School\_Alumni\end{tabular}} & \multicolumn{1}{l|}{\begin{tabular}[c]{@{}l@{}}:almaMater :New\_York\_University\\ :birthPlace :Queens\\ :subject :Tisch\_School\_Alumni\end{tabular}} & \multicolumn{1}{l|}{\begin{tabular}[c]{@{}l@{}}?s :almaMater :New\_York\_University\\ ?s :birthPlace :Queens\\ ?s :subject :Tisch\_School\_Alumni\end{tabular}} \\ \hline
\end{tabular}
}
\caption{Entity Facts, Entity Summary (k=3) \& Entity Summary Pattern}
\label{tab:entity}
\end{table}
\vspace{-1cm}

\subsubsection{Entity Summary Pattern} Given an entity \textit{e} present in an input query triple pattern , i.e. e $\in$  $\left\lbrace \left\langle e, ?p,?o \right\rangle or \left\langle ?s, ?p,e \right\rangle or \ldots \left\langle s, p,e \right\rangle \right\rbrace$, an entity summary pattern is created by picking the  top-\textit{k} ranked facts $f_1,f_2 \ldots f_k$ from the Feature set \textit{FS(e)} i.e. the entity summary. These are in the form of ranked $\langle p,o \rangle$ values where $\left\lbrace f_1,f_2, \ldots, 
f_k \right\rbrace$  = $\left\lbrace \left\langle p_1,o_1 \right\rangle,  \left\langle p_1,o_1 \right\rangle,\ldots \left\langle p_k,o_k \right\rangle \right\rbrace$. To each of these facts a variable \textit{?e} is added to the front such that they become the triple patterns
$\left\lbrace  \left\langle?e,p_1,o_1 \right\rangle , \left\langle ?e,p_2,o_2 \right\rangle, \ldots \left\langle ?e,p_k,o_k\right\rangle \right\rbrace$. The variable \textit{?e} is shared across the triple patterns. \autoref{tab:entity} shows an example of entity fact sets, entity fact summary and an entity summary pattern for the entity :Martin\_Scorsese.


\subsection{Ranking Entity Facts}
Since the aim is to suggest reformulations that give results approximately similar to the original query, the  top-\textit{k} features are ranked by their \textit{specificity} and \textit{popularity}. This is inspired by traditional TF-IDF measures and work in entity summarization\cite{gunaratna2015faces}.

\subsubsection{Specificity}
To pick features that are unique to the entity \textit{e} from \textit{FS(e)} we adapt the well known TF-IDF scheme and model specificity by using the equation below
\begin{equation} \label{eq:1}
Specificity  = \log {\frac{\mid E \mid }{\mid e \mid \exists p,o:\langle e,p,o \rangle \in R\mid }}
\end{equation}

Here $\mid E \mid$ refers to the total number of entity resources in the knowledge graph R. The difference between \cite{gunaratna2015faces} and our work is that we utilize the number of entity resources, rather than all the triple statements in the graph in \ref{eq:1}.  While Equation (\ref{eq:1}) will rank the features in the order of how unique they are, we need to ensure that the features are not too selective and actually offer scope for generating more results when converted to a pattern. To aid in this we define the following equation which calculates the \textit{value frequency} or \textit{popularity} of the entity value. 
\begin{equation} \label{eq:2}
popularity  = \log {\mid t \mid \exists o:\langle e,p,o \rangle \in R\mid }
\end{equation}
In (\ref{eq:2}) we count all the triple statements \textit{t} that have the value \textit{o} in the knowledge graph \textit{G}. Now every feature \textit{f} in the feature set \textit{FS(e)} is ranked by 
\begin{equation} \label{eq:3}
Rank  = Specificity(f) * popularity(o \mid o \in f)
\end{equation}
\subsubsection{Entity Fact Selection}
Due to the nature of extraction, large knowledge graphs like DBpedia properties which are \textit{redundant} or can have statements that refer to values which can satisfy \textit{part-of} conditions. For example the property \textit{birthdate} occurs under both \textit{dbpedia:ontology} and \textit{dbpedia:property}, Similarly an entity can have two statements that have values like (:birthPlace, :Detroit) and (:birthPlace, :Michigan). In this situation it is known that \textit{Detroit} is in \textit{Michigan} but the information cannot be inferred by the triple statements alone. Appending facts such as these is not very informative and would be better if any one is picked. To achieve this we perform the following\footnote{Note that this is only in DBpedia and not for LUBM, and will be generalized in our future work.}:
\begin{itemize}
\item Property Similarity: If there are many properties of the same name and they are in different namespaces, pick the one that is ranked higher.
\item Range Similarity: While picking the top-\textit{k} properties, pick the properties such that their ranges are distinct. This ensures explicit diversity in the property values.
\item Property Groups : Create groups of properties that share the same range. To pick a property from a group for reformulation, pick the highest ranked property. For the second reformulation pick the second highest and so on until there are no more properties and groups.
\item Limit Summary Size : Limit the value of \textit{k} to a heuristic average. In our experiments we found that a value of \textit{k}=3 gave optimal results, This was gleaned from the LSQ dataset\cite{saleem2015lsq} by looking at the size of entity query patterns issued to DBpedia. The highest number of entity queries had input patterns of size ranging from 1 to 3.
\end{itemize}
In addition to this we also ensure that we do not select facts that have \textit{literal values} and facts that are present only once in the entire graph(e.g. unique facts about an entity). We also remove the facts that contain properties, which are not accurate or do not offer any new insight.\footnote{e.g. dbo:viafId. A full list can be found at \url{https://github.com/N00bsie/QueryReformulation/blob/master/README.md}}
\subsection{Query Reformulation Procedure}
Given and input materialized graph \textit{cl}(G) and a query Q containing entity triple patterns and schema elements apply the rules defined in \ref{sec:reformulation} to create reformulated and relaxed queries until no more can be applied or if a stopping condition is predefined. In this paper we only show queries on which \textit{entity reformulation} is performed. However since our work is an extension, these generated reformulations are ordered in a \textit{relaxation graph} as defined in \cite{hurtado2006,fokou2015,fokou2016}. Since our focus is on entity reformulation we utilize the relaxation graphs already created by existing work. Instead of applying relaxation to an entity, we would reformulate using our rules and insert the new query in lieu of the relaxed query at the appropriate levels in the relaxation graph. 

\section{Evaluation}~\label{sec:evaluation}
We conduct two experiments to test our proposed reformulation rules. In the first experiment we evaluate the \textit{scalability} of our proposed entity reformulation system on different sizes of LUBM\cite{guo2005lubm}. In the second, we show how our system generates more contextual results on DBpedia by means of example queries and a discussion of the generated results.

\subsection{Experiment 1} 
\subsubsection{System Setup}: We generate the reformulations on a 64-bit Linux System running on Mint 18.1 based on Ubuntu 16.04.03. The Kernel version is 4.8.0-58. The processor is an Intel\copyright  Core\texttrademark 2.4G CPU with 4 cores and 32 GB RAM. The algorithms were implemented using Apache Jena and the triple store is Jena TDB. 

\begin{table}[htp]
\centering
\resizebox{\columnwidth}{!}{
\begin{tabular}{|l|l|l|}
\hline
   & \multicolumn{1}{c|}{Original Failed Query}                                                                                                                                                                                                            & \multicolumn{1}{c|}{Relaxed Version ( repaired )}                                                                                                                                                                                             \\ \hline
Q1 & \begin{tabular}[c]{@{}l@{}}SELECT,?X ?Y1 ?Y2 ?Y3 ?Y4 WHERE\{ ?X a VisitingProfessor . \\ ?X memberOf \textbf{Department1.University1.edu}.\\ ?X name ?Y1 .\\ ?X emailAddress ?Y2 .\\ ?X telephone ?Y3 .\\ ?X ub:undergraduateDegreeFrom ?Y4 \}\end{tabular}    & \begin{tabular}[c]{@{}l@{}}SELECT,?X ?Y1 ?Y2 ?Y3 ?Y4 WHERE\{ ?X a VisitingProfessor . \\ ?X memberOf ?z.\\ ?X name ?Y1 .\\ ?X emailAddress ?Y2 .\\ ?X telephone ?Y3 .\\ ?X ub:undergraduateDegreeFrom ?Y4 \}\end{tabular}    \\ \hline
Q2 & \begin{tabular}[c]{@{}l@{}}SELECT ?X ?Y  WHERE\{ ?X a Professor . \\ ?X worksFor \textbf{Department0.University0.edu}.\\ ?X researchInterest 'Research2' .?X doctoralDegreeFrom ?Y .\}\end{tabular}                                                            & \begin{tabular}[c]{@{}l@{}}SELECT ?X ?Y WHERE\{ ?X a Professor . \\ ?X worksFor ?Z.\\ ?X researchInterest ?Z1 .\\ ?X doctoralDegreeFrom ?Y .\}\end{tabular}                                                                                   \\ \hline
Q3 & \begin{tabular}[c]{@{}l@{}}SELECT ?X ?Y WHERE\{ \\ \textbf{GraduateStudent73} advisor ?Y . \\ ?Y doctoralDegreeFrom \textbf{University0.edu}.\}\end{tabular}                                                                                                            & \begin{tabular}[c]{@{}l@{}}SELECT ?X ?Y WHERE \{ \\ ?X advisor ?Y . \\ ?Y doctoralDegreeFrom Z.\}\end{tabular}                                                                                                   \\ \hline
Q4 & \begin{tabular}[c]{@{}l@{}}SELECT ?X ?Y1 WHERE\{\\  ?X researchInterest 'Research28'. \\ ?Y1 subOrganizationOf \textbf{University8.edu}.\\ ?X rdf:type Lecturer .\\ ?X worksFor ?Y1 .\}\end{tabular}                                                           & \begin{tabular}[c]{@{}l@{}}SELECT ?X ?Y1 WHERE\{ \\ ?X researchInterest ?Z1. \\ ?Y1 subOrganizationOf ?Z2.\\ ?X rdf:type Faculty .\\ ?X worksFor ?Y1 .\\ \}\end{tabular}                                                          \\ \hline
Q5 & \begin{tabular}[c]{@{}l@{}}SELECT ?X ?Y1?Y2 WHERE \{ \\ ?X a FullProfessor .\\ ?X doctoralDegreeFrom \textbf{University8}.\\ ?X researchInterest 'Research23'. \\ ?X teacherOf ?Y1. \\ ?X takesCourse ?Y1 .\}\end{tabular}                                     & \begin{tabular}[c]{@{}l@{}}SELECT ?X ?Y1?Y2 WHERE \{\\  ?X a FullProfessor .\\ ?X doctoralDegreeFrom ?Z1.\\ ?X researchInterest 'Research23'. \\ ?X teacherOf ?Y1. \\ ?X takesCourse ?Y1 .\}\end{tabular}                             \\ \hline
\end{tabular}
}
\caption{Workload queries}
\label{tab:workload}
\end{table}
\vspace{-1cm}

\subsubsection{Data and Queries} Since our technique extends the query relaxation problem by proposing a reformulation for entity elements, we utilize the  benchmark queries designed by Huang \textit{et al.}\cite{huang2012} as shown in Table~\ref{tab:workload}. These queries were designed to give \textit{zero} results. Huang \textit{et al.} proposed algorithms that relaxed these queries to give additional answers. For our purposes we utilized 5 of the 7 benchmark queries. We picked these queries with the condition that they have entity elements in their triple patterns. We then proceeded to \textit{relax} the entities in the queries and executed these entity relaxed versions. After this we applied our rules to the entities in the queries and executed the generated reformulations. We executed both the relaxations and the reformulations on LUBM10, LUBM20, LUBM50 and LUBM100. The largest of these datasets LUBM100 has about 17M triples. The exact statistics for the dataset sizes can be found in \cite{guo2005lubm}. We now compare and discuss the performances of reformulation against relaxation. 

\vspace{-0.5cm}
 \begin{table}[H]
\centering
\begin{tabular}{|l|l|l|l|l|l|}
\hline
\multicolumn{6}{|c|}{\textbf{Query 1 (Q1)}}                                                                                                                                    \\ \hline
                                      & \multicolumn{1}{c|}{\textbf{Relaxation}} & \multicolumn{1}{c|}{\textbf{Ref-1}} & \textbf{Ref-2} & \textbf{Ref-3} & \textbf{Ref-4} \\ \hline
\multicolumn{1}{|c|}{\textbf{LUBM10}} & 480.4ms/5715                             & 237.2ms/311                          & 311ms/585      & 490.2ms/5715     & 530ms/5715     \\ \hline
\textbf{LUBM50}                       & 1.148s/29976                             & 380.3ms/1529                        & 288ms/585       & 1.219s/29976    & 1.576ms/29976    \\ \hline
\textbf{LUBM100}                      & 1.981s/60268                             & 315ms/3023                          & 250ms/585        & 1.910s/60268   & 1.920s/60268   \\ \hline
\end{tabular}
\caption{Query 1 results (execution times/number of results)}
\label{tab:query1}
\end{table}
\vspace{-1.25cm}
\begin{table}[H]
\centering
\begin{tabular}{|l|l|l|l|l|l|}
\hline
\multicolumn{6}{|c|}{\textbf{Query 2 (Q2)}}                                                                                                                                    \\ \hline
                                      & \multicolumn{1}{c|}{\textbf{Relaxation}} & \multicolumn{1}{c|}{\textbf{Ref-1}} & \textbf{Ref-2} & \textbf{Ref-3} & \textbf{Ref-4} \\ \hline
\multicolumn{1}{|c|}{\textbf{LUBM10}} & 325ms/5715                               & 192ms/297                              & 250ms/447       & 250ms/447       & 380ms/5715      \\ \hline
\textbf{LUBM50}                       & 1.050s/29976                             & 352.3ms/1480                        & 358ms/447      & 250ms/447       & 872ms/29976    \\ \hline
\textbf{LUBM100}                      & 1.081s/60268                             & 471.5ms/2980                          & 512ms/447      & 316ms/447      & 1.502s/60268     \\ \hline
\end{tabular}
\caption{Query 2 results (execution time/ number of results)}
\label{tab:query2}
\end{table}
\vspace{-1.5cm}
\begin{table}[H]
\centering
\begin{tabular}{|c|l|l|l|l|l|}
\hline
\multicolumn{6}{|c|}{\textbf{Query 3  (Q3)}}                                                                                                     \\ \hline
\multicolumn{1}{|l|}{} & \multicolumn{1}{c|}{\textbf{Relaxation}} & \multicolumn{1}{c|}{\textbf{Ref-1}} & \textbf{Ref-2} & \textbf{Ref-3} & \textbf{Ref-4} \\ \hline
\textbf{LUBM10}        & 145ms/0                                  & 130ms/0                             & 145ms/0        & 168ms/0        & 175ms/0        \\ \hline
\end{tabular}
\caption{Query 3(single entity reformulated) results on LUBM10 (execution time/ number of results}
\label{tab:query3.1}
\end{table}
\vspace{-1cm}
\begin{table}[H]
\centering
\begin{tabular}{|l|l|l|l|l|l|}
\hline
\multicolumn{6}{|c|}{\textbf{Query 3  (Q3)}}                                                                                                                                       \\ \hline
                                      & \multicolumn{1}{c|}{\textbf{Relaxation\#2}} & \multicolumn{1}{c|}{\textbf{Ref-1}} & \textbf{Ref-2} & \textbf{Ref-3} & \textbf{Ref-4} \\ \hline
\multicolumn{1}{|c|}{\textbf{LUBM10}} & 541ms/39255                                 & 160ms/0                             & 139ms/8        & 152ms/6        & 204ms/255      \\ \hline
\textbf{LUBM50}                       & 3.76s/204930                                & 305ms/0                             & 274ms/8        & 267ms/6        & 463/255        \\ \hline
\textbf{LUBM100}                      & 9.5s/411501                                 & 347ms/0                             & 448ms/8        & 400ms/6        & 703ms/255      \\ \hline
\end{tabular}
\caption{Query 3(both entities reformulated) results (execution time/number of results)}
\label{tab:query3.2}
\end{table}
\vspace{-1cm}

\begin{table}[htp]
\centering
\scalebox{0.85}{
\begin{tabular}{|l|l|}
\hline
\multicolumn{1}{|c|}{\textbf{Department1.University1.edu}}                                                                                          & \multicolumn{1}{c|}{\textbf{Department0.University0.edu}}                                                                                      \\ \hline
\begin{tabular}[c]{@{}l@{}}name  'Department1'\\ subOrganizationOf  University1.edu\\ type  Department\\ type  Organization\end{tabular} & \begin{tabular}[c]{@{}l@{}}name 'Department0'\\ type Institute \\ subOrganizationOf  University0.edu\\ type Department\\ type Organization\end{tabular} \\ \hline
\textbf{University8}                                                                                                                                & \textbf{GraduateCourse65}                                                                                                                      \\ \hline
\begin{tabular}[c]{@{}l@{}}type University\\ type Organization\end{tabular}                                                                & \begin{tabular}[c]{@{}l@{}}name GraduateCourse65\\ type  GraduateCourse\\ type Course \\ type Work\end{tabular}                       \\ \hline
\end{tabular}
}
\caption{Instances \& Ranked Features}
\label{tab:rankedLUBMFeatures}
\end{table}

\subsubsection{Reformulation Performance} The workload queries are shown in \autoref{tab:workload} with the entities in bold. For these entities in the benchmark queries, we first generate the ranked features for the entities. These are shown in \autoref{tab:rankedLUBMFeatures}. The calculation time for these features on LUBM datasets is negligible. Since the total number of properties in LUBM for entities are between 5 and 7, We create one new reformulation per feature to the original query as per the rules defined in \autoref{tab:rules}. Heuristically appending more than one reformulation triple in LUBM makes the new query highly \textit{selective} and so we append each feature one by one to the original query. In this case, if there are \textit{m} features, we generate \textit{m} reformulations. Table~\ref{tab:query1} and Table~\ref{tab:query2} show the performance of the first two workload queries with four reformulations along with the time taken to execute and the number of results generated. Huang~\textit{et. al}~\cite{huang2012} evaluated their relaxations by computing the number of relaxations that would generate a certain number of answers \textit{K}, with \textit{K}=(10,50,150). In such a scenario \textbf{Ref-1} of Q1 can be used because it gives a more precise set of answers for roughly a lesser execution time on LUBM10, LUBM50, LUBM100. 

In contrast, the relaxation gives a generalized set of results. It is also seen that \textbf{Ref-4} of Q1 generates the same number of answers as the \textit{relaxation}. However, despite the same number of answers, it is worthwhile to note that \textbf{Ref-4} is contextualized by the addition of a pattern that binds the type of the variable. This is same Q2 as well. The workload query in Q3 has two entities; without performing any checks, if we reformulate just the first entity pattern, we end up with Table~\ref{tab:query3.1}, which shows zero results. After this, we reformulate the second entity pattern in Q3 as well. The results of these are shown in Table~\ref{tab:query3.2}. As shown by these reformulations, the number of additional results are low. This is because the feature patterns have high selectivity, but we note that they have more relevance. 
Our implementation can be found at \href{https://github.com/N00bsie/QueryReformulation}{our GitHub.}
\begin{table}[H]
\centering
\resizebox{0.85\columnwidth}{!}{
\begin{tabular}{|l|l|l|}
\hline
                                  & \multicolumn{1}{c|}{\textbf{Q}}                                                                                                    & \multicolumn{1}{c|}{\textbf{Q$^{\prime}$}}                                                                                                               \\ \hline
\multicolumn{1}{|c|}{} & \begin{tabular}[c]{@{}l@{}}SELECT DISTINCT ?movie \\ WHERE \{\\ ?movie dbo:director dbr:Francis\_Ford\_Coppola .\\ \}\end{tabular} & \begin{tabular}[c]{@{}l@{}}SELECT DISTINCT ?movie \\ WHERE \{\\ ?movie a dbo:Film .\\ ?movie dbo:director dbr:Francis\_Ford\_Coppola .\\ \}\end{tabular} \\ \hline
\end{tabular}
}
\caption{QALD-2 Query}
\label{tab:DBpedia-input}
\end{table}

\subsection{Applications to real-world data}
In order to demonstrate the applicability of our work in real-world situations, we applied our approach on the large knowledge graph DBpedia. Specifically, we gathered 20 queries from the QALD-2 task challenge\footnote{\url{https://qald.sebastianwalter.org/2/data/dbpedia-test.xml}} with associated ground truth results. These 20 queries were chosen such that they result in lists of entities and have entity patterns in them. Though the task in QALD-2 is about translating natural language questions to SPARQL, it also provides test SPARQL annotations which we used as an input query for our technique. We then compare and contrast the results that are achieved by \textit{relaxation} on the query. For example, for a query \textit{Give me all the movies directed by Francis Ford Coppola}, the initial query \textbf{Q} is shown in \autoref{tab:DBpedia-input}.  By inferring the type of the variable \textit{?movie}, it is assigned a type \textit{dbo:Film} from the list of available types. When we execute this reformulation on the input graph \href{http://downloads.dbpedia.org/current/}{DBpedia 2015-10}, we get a total of 32 results whose values are the movies directed by \textit{Francis Ford Coppola}. If a user is not satisfied with these 32 answers and is looking for additional answers then the relaxed versions of the query are Q1$^{\prime}$ and Q2$^{\prime}$. 

\vspace{-0.5cm}
\begin{table}[H]
\centering
\resizebox{1\columnwidth}{!}{
\begin{tabular}{|l|l|l|l|}
\hline
                                                                                                 & \multicolumn{1}{c|}{\textbf{Q1}}                                                                                                                                           & \multicolumn{1}{c|}{\textbf{Q1$^{\prime}$}}                                                                                                                                & \multicolumn{1}{c|}{\textbf{Q2$^{\prime}$}}                                                                                                                                         \\ \hline
\multicolumn{1}{|c|}{\textbf{Query}}                                                             & \textbf{\begin{tabular}[c]{@{}l@{}}SELECT DISTINCT ?movie where\\  \{\\ ?movie a dbo:Film .\\ ?movie dbo:director dbr:Francis\_Ford\_Coppola .\\ \}\end{tabular}}          & \textbf{\begin{tabular}[c]{@{}l@{}}SELECT DISTINCT ?movie where \\ \{\\ ?movie a dbo:Film .\\ ?movie ?y .\\ \}\end{tabular}}          & \textbf{\begin{tabular}[c]{@{}l@{}}SELECT DISTINCT ?movie where \\ \{\\ ?movie a dbo:Work .\\ ?movie dbo:director ?y .\\ \}\end{tabular}}                                           \\ \hline
\textbf{Results/Time}                                                                            & \multicolumn{1}{c|}{\textbf{32/4ms}}                                                                                                                                       & \multicolumn{1}{c|}{\textbf{83776/1.77s}}                                                                                                                                      & \multicolumn{1}{c|}{\textbf{102577/2.19s}}                                                                                                                                          \\ \hline
\multicolumn{1}{|c|}{\textbf{\begin{tabular}[c]{@{}c@{}}Result Snippet\\ size = 5\end{tabular}}} & \begin{tabular}[c]{@{}l@{}}dbr:Tetro\\ dbr:Peggy\_Sue\_Got\_Married\\ dbr:The\_Godfather\_Saga\\ dbr:The\_Terror\_(1963\_film)\\ dbr:Battle\_Beyond\_the\_Sun\end{tabular} & \begin{tabular}[c]{@{}l@{}}dbr:12\_Monkeys\\ dbr:American\_Beauty\\ dbr:Amélie\\ dbr:Animal\_Crackers\_(1930\_film)\\ dbr:Anne\_of\_the\_Thousand\_Days\end{tabular} & \begin{tabular}[c]{@{}l@{}}dbr:...All\_the\_Marbles\\ dbr:20,000\_Years\_in\_Sing\_Sing\\ dbr:5ive\_days\_to\_Midnight\\ dbr:881\_(film)\\ dbr:A\_Home\_on\_the\_Range\end{tabular} \\ \hline
\end{tabular}
}
\caption{Relaxation Summary(number of results, execution time)}
\label{tab:dbpedia-rel}
\end{table}
\vspace{-1cm}

Assuming the user is interested in more movies, relaxation changes the instance \textit{Francis\_Ford\_Coppola} to a variable \textit{?y}. This results in 83776 answers for Q1$^{\prime}$, and a snippet of five results is shown in Table~\ref{tab:dbpedia-ref}.  
Q2$^{\prime}$ results in 102577 answers because \textit{dbo:Film} is relaxed to \textit{dbo:Work}. Now the new relaxation not only includes movies but also includes entities of type \textit{dbo:MusicalWork}. Not all of them would be relevant to the user and the onus is again on the user to search for answers that would suit his information needs. 
\begin{table}[H]
\centering
\resizebox{0.95\columnwidth}{!}{
\begin{tabular}{|l|l|l|}
\hline
                                                                                                             & \multicolumn{1}{c|}{\textbf{R1$^{\prime}$}}                                                                                                                                                                                                & \multicolumn{1}{c|}{\textbf{R2$^{\prime}$}}                                                                                                                                                                                     \\ \hline
\multicolumn{1}{|c|}{\textbf{Query}}                                                                         & \textbf{\begin{tabular}[c]{@{}l@{}}SELECT DISTINCT ?movie where\\  \{\\ ?movie a dbo:Film .\\ ?movie dbo:director ?z .\\ ?z a wiki:Q215627 .\\ ?z dbo:birthPlace dbr:Detroit .\\ ?z dbo:occupation dbr:Film\_Producer .\\ \}\end{tabular}} & \textbf{\begin{tabular}[c]{@{}l@{}}SELECT DISTINCT ?movie where \{\\ ?movie a dbo:Film .\\ ?movie dbo:director ?z.\\ ?z a wiki:Q5\\ ?z dbo:birthPlace dbr:Michigan .\\ ?z dbo:occupation dbr:Screenwriter . \\ \}\end{tabular}} \\ \hline
\textbf{Toal Results/Time}                                                                                   & \multicolumn{1}{c|}{\textbf{62/12ms}}                                                                                                                                                                                                      & \multicolumn{1}{c|}{\textbf{100/20ms ( 38 additional)}}                                                                                                                                                                         \\ \hline
\multicolumn{1}{|c|}{\textbf{\begin{tabular}[c]{@{}c@{}}Results other\\ than Relaxed versions\end{tabular}}} & \begin{tabular}[c]{@{}l@{}}dbr:Art\_History\_(film)\\ dbr:Caitlin\_Stainken\\ dbr:Indian\_Summer\\ dbr:Silver\_Bullets\\ dbr:Man\_About\_Town\_(2006\_film)\end{tabular}                                                                   & \begin{tabular}[c]{@{}l@{}}dbr:Getting\_Gilliam\\ dbr:Hannah\_Takes\_the\_Stairs\\ dbr:Narc\_(film)\\ dbr:Elevated\_(film)\\ dbr:Cypher\end{tabular}                                                                            \\ \hline
\end{tabular}
}
\caption{Reformulations and Results}
\label{tab:dbpedia-ref}
\end{table}
\vspace{-1cm}

\begin{table}[H]
\centering
\resizebox{0.75\columnwidth}{!}{
\begin{tabular}{|l|l|l|}
\hline
                                  & \multicolumn{1}{c|}{\textbf{R1$^{\prime}$}}                                                                                                               & \multicolumn{1}{c|}{\textbf{R2$^{\prime}$}}                                                                                                                                                                \\ \hline
\multicolumn{1}{|c|}{\textbf{Related Entities}} & \begin{tabular}[c]{@{}l@{}}dbr:Francis\_Fod\_Coppola\\ dbr:Joe\_Swanberg\\ dbr:Tariq\_Nasheed\\ dbr:Mike\_Binder\\ dbr:Berry\_Gordy\end{tabular} & \begin{tabular}[c]{@{}l@{}}dbr:Francis\_Ford\_Coppola\\ dbr:Vincenzo\_Natali\\ dbr:Joe\_Swanberg\\ dbr:Joe\_Carnahan\\ dbr:Mike\_Binder\\ dbr:Paul\_Schrader\\ dbr:Oren\_W.\_Haglund\end{tabular} \\ \hline
\end{tabular}
}
\caption{Bounded Variable values for Reformulation}
\label{tab:bounded}
\end{table}
\vspace{-0.5cm}
Now lets reformulate the the entity \textit{Francis\_Ford\_Coppola} with summary patterns of size \textit{k}=3.  We get the queries R1$^{\prime}$ and R2$^{\prime}$ shown in \autoref{tab:dbpedia-ref}. The summary pattern features are shown and they are all bounded to a variable \textit{?z}, whose   values(\autoref{tab:bounded}) are more \textit{similar} to \textit{Francis\_Ford\_Coppola} than the relaxed variable \textit{?y}. This is because in R1$^{\prime}$ the entities are related by \textit{birthPlace, Detroit} and \textit{occupation,Film Producer} and in R2$^{\prime}$ they are related by \textit{birthPlace, Michigan} and \textit{occupation, Screen Writer}. Thus the generated results are more contextual to the original query than in relaxation. In addition the number of results is more manageable and the query execution time is lesser than the relaxations.  

\section{Discussion}\label{sec:discussion}
Query reformulation or Query rewriting has been an active topic of research spanning relational databases, text search systems, logical databases and now graph databases.While text search systems have to use heuristics to perform reformulations, relational databases have relied on the database schema to suggest reformulations. Most of the work on relaxation is built upon the seminal work of  Gaasterland et al. \cite{gaasterland1992}. Instead of using the term rewriting of queries, they proposed the term \textit{relaxation} for deductive databases and logic programming queries. The authors argued that Database answers to a query, though logically right, can sometimes be misleading. So it would be of interest to present additional answers achieved by relaxing the constraints on the query. They also surmised that having a taxonomy would ideally be helpful to relax conditions. 
Thus \textit{Query Relaxations} are viewed as a mechanism where the user query is relaxed to include more relevant additional automatic queries.The first techniques relied on generalizing the query triple patterns and using class and property hierarchies was proposed by Hurtado \textit{et al.}\cite{hurtado2008}. They suggested the use of the RELAX operator on triple patterns to enable \textit{flexible querying}. \cite{huang2012} used the rules proposed by \cite{hurtado2006} and devised automatic relaxations instead of applying a RELAX clause. These relaxations were then applied to \textit{failed queries} i.e. queries which gave a zero answer. \cite{cali2014flexible} also propose new operators APPROX and RELAX to tackle the notion of \textit{flexible} querying. In addition to RDF based techniques, there are other systems which view at relaxation and reformulation as aiding the original user intent.  These techniques produce relaxations by combining statistical Language Models and structured querying over RDF data \cite{elbassouni2011}. Relaxation has also been used as a mechanism to find approximate answers for fuzzy queries on crisp data\cite{hogan2012}.  A recent work on \textit{cooperative answering} has also used the notion of Godfrey's failing subqueries \cite{godfrey1997minimization} by finding \textit{minimal failing queries} \cite{fokou2015}. They then extended this work to include relaxations by applying relaxations only to elements present in the \textit{minimal failing queries}. 

While such systems work on the concept and relationship level and on only schema elements, none of them have proposed to reformulate the \textit{entities or resources} in the query. We believe that our work on entity reformulation will work in conjunction with existing techniques to create a robust query reformulation system. In addition our technique is extensible to not just AND patterns, but also UNION and other graph query patterns. This will be explored in future work.

\section{Conclusion and Future Work}\label{sec:conclusion}
In this paper, we have proposed a query reformulation strategy based on  the unique properties that define entities in knowledge graphs.  We have shown that though semantic queries can often be crisp , there is a need for efficient and effective reformulation in situations where entity patterns are present. We observed that there are several strategies for this reformulation---especially in past literature---but they typically resulted in generalizations. Our approach based on entity centric reformulation extends this and utilizes schema information and entity features present in the graph to suggest rewrites. This has shown promising results as discussed in the results section. To minimize the complexity of reformulations w.r.t. these features, we introduce strategies that select the top-k most \emph{relevant} and \emph{informative} features and use them to augment the original query to create a valid graph query. In the future we will work on \emph{ranking mechanisms} to generate more relevant reformulations. Another challenge is to devise strategies for the order in which \emph{relaxation} and \emph{reformulation} should be performed if there are both schema and entity elements in a pattern. We hope to address these issues in near future and present our findings in a future venue.
%
%
\bibliographystyle{splncs03}

\end{document}